\documentclass[]{bluepixel}

\usepackage{microtype}
\usepackage{graphicx}
\usepackage{subcaption}
\usepackage{ragged2e}
\DeclareCaptionJustification{RaggedRightHyphenated}{\RaggedRight}
\usepackage{csquotes}
\usepackage{afterpage}
\usepackage{float}
\usepackage{fontawesome5}

\usepackage{amsmath}
\usepackage{amssymb}
\usepackage{mathtools}
\usepackage{amsthm}
\usepackage{xspace}
\usepackage{colortbl}
\usepackage{multirow}
\usepackage{enumitem}
\usepackage{wrapfig}
\usepackage{nicefrac}

\newcommand{\bpxmodel}{BluePixel\xspace}

\definecolor{bpxApp}{HTML}{c8e7fa}
\definecolor{bpxPurple3}{HTML}{f0ebf5}

\definecolor{citecolor}{HTML}{2563EB}
\definecolor{linkcolor}{HTML}{2563EB}

\title{BluePixel: Toward Native 4K Agentic Image Editing}

\author[1]{Shanyan Guan}
\author[1]{Shiyu Li}
\author[1]{Xin Li}
\author[2]{Zhizhou Chen}
\author[2]{Yunzhe Xu}
\author[3]{Yuehao Liu}
\author[1]{Yifan Lu}
\author[1]{Ruixue Shen}
\author[1]{Shiju Zhang}
\author[1]{Min Jin}
\author[1]{Qiang Liu}
\author[1]{Junbo Liu}
\author[1]{Penggang Shi}
\author[1]{Sheng Sun}
\author[1]{Xiao Wang}
\author[1]{Yong Zhang}
\author[1]{Zhongang Qi}
\author[1]{Wei Li}
\author[3]{Chao Ma}
\author[2]{Ying Tai}
\author[1]{Yanhao Ge}

\affiliation[1]{vivo Mobile Communication Co., Ltd.}
\affiliation[2]{Nanjing University}
\affiliation[3]{Shanghai Jiao Tong University}

\abstract{
Instruction-driven image editing has reached native 4K, yet how such editors are trained remains undisclosed, and pretraining one from scratch is beyond most teams. The task itself compounds the difficulty: a single end-to-end editor cannot make precise parametric adjustments, match specialized experts, or interpret the vague instructions users actually write. \textbf{We argue that 4K editing is a system problem rather than a single-model problem, and present \bpxmodel, which couples a native-4K editing model with an agent harness; the complete training recipe is disclosed in this paper.} The model makes 4K training practical: it turns a 1K-pretrained editor into a native-4K one through progressive VAE adaptation that keeps 4K training as affordable as 1K, absorbs a long tail of vertical editing skills through continual learning, and refines the aesthetic--fidelity trade-off with reinforcement learning. Distillation via \emph{source-anchored target projection} (SATP) then compresses 50 denoising steps to 4, \textbf{completing a 4K edit in 8.7\,s on a single Kunlunxin P800 XPU} while preserving task success rate and background consistency. The harness, BluePixel Agent, makes editing controllable: it decomposes user instructions into specialist and general operations, verifies every result with the same reward model that guided training, and retries within bounded attempts. On VINS-4KEval++ (3{,}305 native-4K cases), \bpxmodel ranks second among seven models, behind only GPT Image 2 in non-edited-region consistency.
}
\metadata[Project Page]{\faGlobe\ \url{https://tech.vivo.com/BluePixel}}
\metadata[Code]{\faGithub\ \url{https://github.com/vivo/BluePixel}}
\metadata[Model]{\faRobot\ \url{https://huggingface.co/openvivo} \quad \faCubes\ \url{https://modelscope.cn/organization/vivo}}

\begin{document}

\maketitle

%% ================================================================
%%  INTRODUCTION
%% ================================================================

\section{Introduction}
\label{section:intro}

\begin{figure}[tp]
    \centering
    \includegraphics[width=\linewidth]{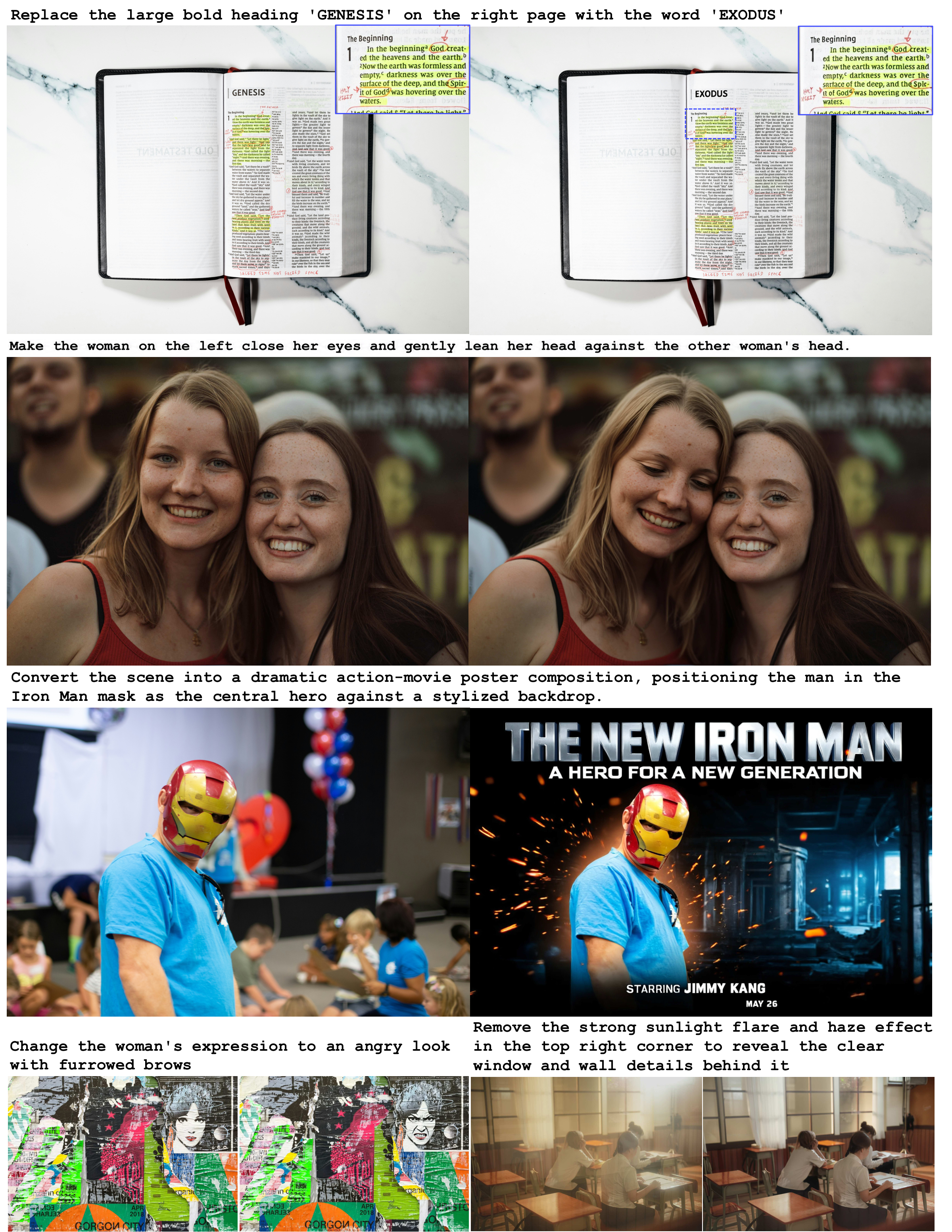}
    \caption{\textbf{\bpxmodel editing showcase.} Diverse 4K editing capabilities including text editing, pose editing, scene recomposition, anime-style expression editing, and low-level image editing. The model preserves fine details faithfully: even small text remains crisp and legible.}
    \label{fig:elo_teaser}
\end{figure}

Diffusion models~\citep{rombach2022high, podell2023sdxl, esser2024scaling} have brought photorealistic image synthesis and instruction-driven editing~\citep{brooks2023instructpix2pix, kawar2023imagic} to production scale, and output resolution keeps rising: closed-source models went from 1K to 2K and then to 4K, and editing has now reached native 4K as well. A straightforward idea is to cascade a 1K editor with a super-resolution or repainting model~\citep{wang2021real, liang2021swinir}; the cascade, however, introduces structural artifacts at the upsampling boundary (ghosting, texture repetition, and loss of fine-grained control) that professional workflows cannot accept, as shown in~\citep{guan2026vins}. VINS-120K~\citep{guan2026vins} first proposed post-training a 1K-pretrained editor directly into a native-4K editing model, and introduced an ultra-high-resolution editing dataset to make this possible. More recently, open-source releases such as FLUX.3 Image~\citep{flux3image} and SenseNova-U1.5~\citep{sensenovau15} have followed the move to 4K, but their training recipes remain undisclosed. This paper starts from two questions. How can an efficient native-4K editing model be trained with limited resources? And is editing fundamentally a single-model problem, or should it be solved by an agent framework?

The second question goes deeper than resolution. The dominant paradigm treats editing as a single end-to-end model: one network takes an image and an instruction and returns the edited result~\citep{brooks2023instructpix2pix, sheynin2024emu, meng2021sdedit}. This formulation conflates heterogeneous capabilities into one set of weights, and in practice it falls short in three ways. It cannot make precise, parametric adjustments: an exact brightness step or a specific color grade comes back only as an approximation. On task types such as text rendering, teeth retouching, and hairstyle editing, it trails dedicated expert models. And it offers no principled way to act on the short, vague instructions that users actually write (``make this photo look more professional'').

We answer both questions with BluePixel, whose complete training recipe is disclosed in this paper; \Cref{fig:elo_teaser} previews its editing capabilities. First, the training framework is efficient: it builds on a 1K-pretrained editor and never pretrains at 4K. A progressive VAE adaptation recipe completes the jump to native 4K (4K-in, 4K-out, no cascaded super-resolution) in four days on 32 Kunlunxin P800 XPUs, continual learning with ACE-LoRA~\citep{acelora2026} and MOPD~\citep{diffusionopd2026} absorbs a long tail of vertical editing skills, and reinforcement learning refines the aesthetic--fidelity trade-off. Second, the resulting model is fast: the high-compression VAE keeps the DiT-stage token count of a native-4K edit at the level of a 1K edit, and distillation via \emph{source-anchored target projection} (SATP) then compresses 50 denoising steps to 4, enough for interactive multi-turn editing. Third, we treat editing as an agent framework rather than a single model: BluePixel Agent separates intent analysis from editing, composes specialist and general-purpose tools, and uses reward-based feedback to retry or select among candidate images; persisted image provenance supports multi-turn refinement. A data pipeline (a VLM-based quality gate and a four-dimensional reward model) holds all three to the same quality standard.

The remainder of this paper is organized as follows. \Cref{section:related_work} surveys related work. \Cref{section:bluepixel} presents \bpxmodel, covering both the native 4K training recipe and the agentic editing harness. \Cref{section:data_pipeline} presents data construction and quality assessment. \Cref{section:experiments} presents experiments, and \Cref{section:conclusion} concludes.

%% ================================================================
%%  RELATED WORK
%% ================================================================

\section{Related Work}
\label{section:related_work}

\paragraph{Instruction-driven image editing.}
InstructPix2Pix~\citep{brooks2023instructpix2pix} established instruction-conditioned editing, and subsequent work improved control through test-time optimization~\citep{kawar2023imagic,meng2021sdedit}, larger training data~\citep{sheynin2024emu}, posterior sampling~\citep{tian2025postedit}, hybrid prompt inversion~\citep{guan2024hybridbooth}, and semantic intent understanding~\citep{guan2025describe,guan2026swapping}. Industry models such as HiDream-I1~\citep{hidreami1} have since carried instruction editing to production scale, and their resolution is climbing accordingly: Seedream~4.0/5.0~\citep{seedream4,seedream5}, Qwen-Image~\citep{qwenimage}, and ChatGPT-Image~\citep{gptimage} already generate 4K images; recent open-source releases such as FLUX.3 Image~\citep{flux3image} and SenseNova-U1.5~\citep{sensenovau15} have followed suit, though their training recipes remain undisclosed. The academic line, meanwhile, still operates at $1024\times1024$ or below, and all of these editors keep the same shape: a single end-to-end network that takes an image and an instruction and returns the result. BluePixel keeps this interface and lifts it to native 4K, with its training recipe fully described in \Cref{subsection:4k_training}.

\paragraph{Training at native 4K.}
Scaling training to native 4K poses two resource problems. On the data side, UltraHR-100K~\citep{zhao2025ultrahr} and VINS-120K~\citep{guan2026vins} provide large-scale high-resolution editing pairs, and learnable posterior sampling~\citep{bao2026learnir} advances high-resolution restoration. On the architecture side, a standard $f{=}8$ VAE turns a $4096\times4096$ frame into a latent sequence that makes training intractable, and the common alternative, cascading a 1K editor with super-resolution~\citep{wang2021real,liang2021swinir}, avoids that training cost but introduces structural artifacts (ghosting, texture repetition, and loss of fine-grained control). BluePixel resolves the architecture half: it replaces the standard VAE with DCAE-F64~\citep{dcae} ($f{=}64$ vs.\ $f{=}8$), cutting the 4K token count by $16\times$, to the level of a 1K edit, so that fine-tuning becomes tractable, and it targets \emph{editing} rather than generation, which additionally demands instruction fidelity and non-edited-region consistency.

\paragraph{Agentic image editing.}
Agentic systems delegate task steps to tools under a language-model orchestrator~\citep{compoundai2024}, a paradigm established in text and code~\citep{schick2024toolformer,shen2023hugginggpt,yao2023react,park2023generative,wang2024voyager} and recently extended to visual creation and editing~\citep{divideconquer2024,mulan2024,diffusionagent2024,cca2024,costa2025,fasta2025,mira2025,llmlvlm2025,crea2025,cameo2026,imageeditr1_2026,agentbanana2026}. BluePixel Agent follows this line but changes what the loop verifies and what the tools are: turn-level routing and persistent image provenance are coupled to a structured editing pipeline that delegates text, tone, character effects, and general edits to different tools, and a reward model judges every candidate, driving local retries and final selection rather than serving only as a post hoc quality report.

\section{BluePixel}
\label{section:bluepixel}

In this paper, we aim to address two challenges: making native-4K editing \emph{practical} (efficient training and fast inference with limited resources), and making editing more \emph{controllable} than a single model can be (structured tool use and feedback). \bpxmodel is a native 4K agentic image editing model built around these two goals. Its training recipe (\Cref{subsection:4k_training}) turns a 1K-pretrained editor into a native-4K editor with 4-step inference, and its agentic editing harness (\Cref{subsection:harness}) combines routing, composable editing tools, reward-guided retries, and candidate selection.

\subsection{Efficient Native 4K Training}
\label{subsection:4k_training}

\paragraph{Overview.}
\label{subsubsection:training_overview}

Native 4K image editing poses two structural challenges. \emph{First, token explosion:} a standard $f{=}8$ VAE compresses a $4096{\times}4096$ frame into a latent sequence so long that training and inference become prohibitively expensive for diffusion transformers (DiTs). \emph{Second, data scarcity:} high-quality 4K editing pairs are rare and costly to produce, so the data volume needed to train a 4K model from scratch is difficult to assemble. \bpxmodel addresses both with a \emph{three-phase} training recipe. The recipe is general: it upgrades any pretrained low-resolution editor into a native 4K editor and does not assume a particular base model. For concreteness, we present it on FLUX.2 [klein]~\citep{flux2klein}, the 1K-pretrained editor we build on:

\begin{itemize}\itemsep0pt
    \item \textbf{Phase 1: Preparation.} Replace the $f{=}8$ VAE with DCAE-F64~\citep{dcae} ($f{=}64$) and adapt the DiT via representation alignment plus light fine-tuning to obtain an initial 4K model at low cost ($\sim$200K samples, 32 Kunlunxin P800 XPUs for 4 days).
    \item \textbf{Phase 2: Enhancement.} Progressively improve the model through (i) progressive quality supervised fine-tuning (SFT) at $1024{\times}1024$ then $4096{\times}4096$, (ii) benchmark-driven fine-tuning with data rebalancing and model merging, (iii) continual learning~\citep{acelora2026} for vertical editing skills, and (iv) reinforcement learning with DiffusionNFT~\citep{zheng2025diffusionnft} for aesthetics and instruction fidelity.
    \item \textbf{Phase 3: Acceleration.} Compress the 50-step teacher into a 4-step student via \emph{source-anchored target projection} (SATP) step distillation for production latency.
\end{itemize}

\subsubsection{Phase 1: Preparation --- Lightweight 4K Initialization}
\label{subsubsection:preparation}

The first bottleneck at 4K is the VAE. The native VAE of FLUX.2 [klein] has a down-sampling factor of $f{=}8$ with a $2{\times}2$ patchify; for a 4K frame this produces a latent sequence far beyond what a diffusion transformer can attend over in bf16 precision. Inspired by DC-Gen~\citep{dcgen2025}, which accelerates diffusion models by post-training in a deeply compressed latent space, we replace the native VAE with DCAE-F64~\citep{dcae} ($f{=}64$, $1{\times}1$ patchify). The token count for a $H{\times}W$ frame is
\begin{equation}
    N_{\text{tokens}} = \frac{H}{f \cdot p} \times \frac{W}{f \cdot p},
    \label{eq:token_count}
\end{equation}
so, per \Cref{eq:token_count}, moving from $f{=}8$ to $f{=}64$ reduces the token count of a $4096{\times}4096$ frame by $16\times$, to exactly the token count of a $1024{\times}1024$ edit under the native VAE --- 4K fine-tuning thus fits the same memory footprint as 1K training.

\begin{figure}[!t]
    \centering
    \includegraphics[width=\linewidth]{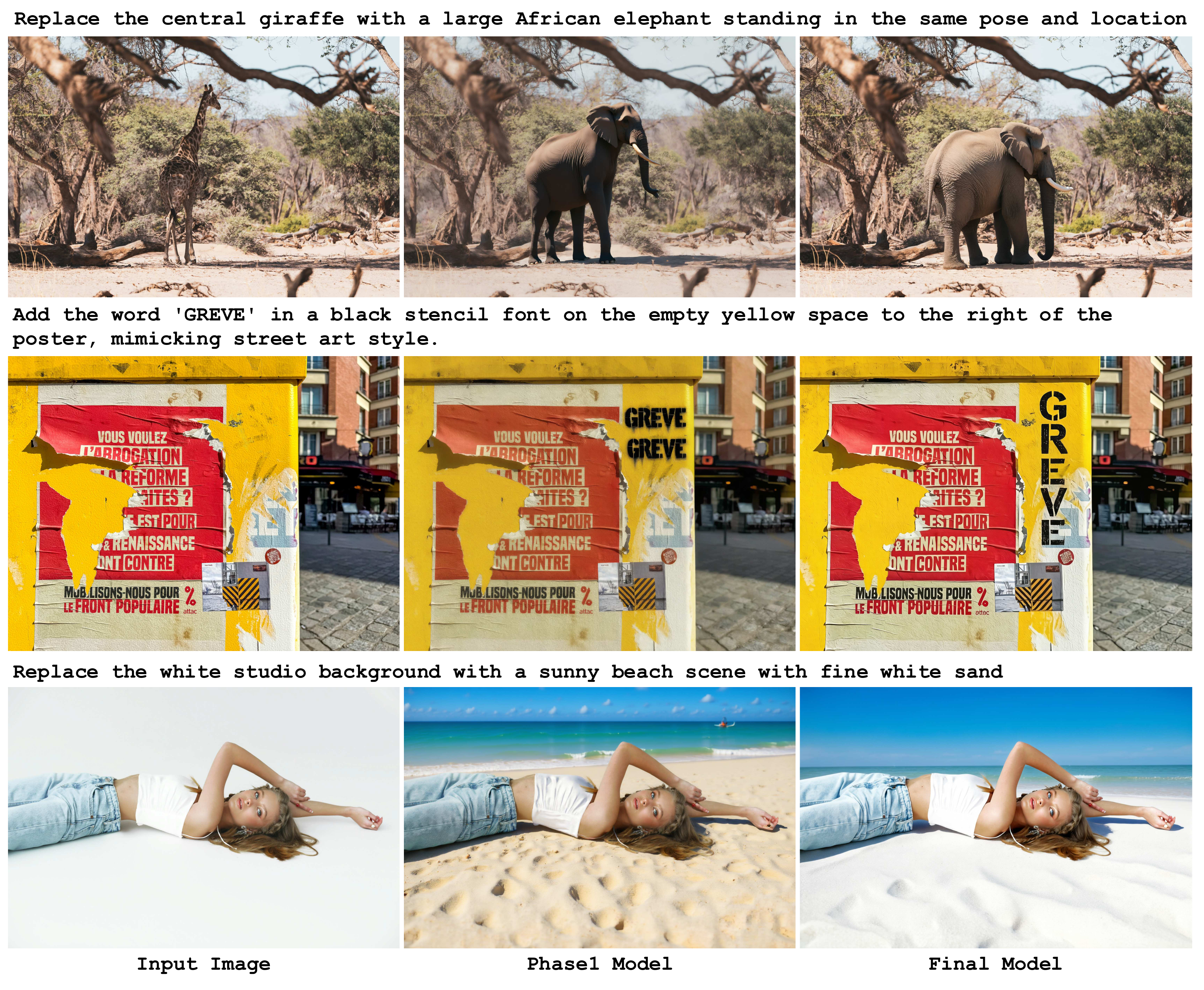}
    \caption{\textbf{Phase 1 vs.\ final model.} The Phase 1 checkpoint produces usable but limited results; after Phase 2 enhancement (progressive SFT, benchmark-driven fine-tuning, continual learning, and reinforcement learning), the final model achieves substantially higher editing quality and detail fidelity.}
    \label{fig:phase1_vs_final}
\end{figure}

However, the public DCAE-F64 latent has 128 channels (vs.\ 32 in the native VAE), so the DiT's input projection is incompatible. The natural remedy is to retrain this projection for the 128-channel latent and continue training, but we find that adapting directly from $f{=}8$ to $f{=}64$ is infeasible: the compression ratio is too aggressive for a single step. Instead, we adopt a \emph{progressive} adaptation through $f\!:\;8 \to 32 \to 64$, where each step bridges only a moderate representation gap. At each compression factor, we perform a three-step adaptation: (i) \emph{input projection distillation}: distill a new input projection layer via dual-VAE alignment,
\begin{equation}
    \mathcal{L}_{\text{align}} = \mathbb{E}\!\left[\big\| g_\theta\!\big(E_{\text{new}}(\mathbf{x}),\, t\big) - g_\phi\!\big(E_{\text{old}}(\mathbf{x}),\, t\big) \big\|_2^2\right],
    \label{eq:align_loss}
\end{equation}
where $g_\theta$ and $g_\phi$ map latents from the new and original VAEs (with the timestep $t$) into the DiT's input space; $g_\phi$ is frozen and only $g_\theta$ is trained, so the DiT receives consistent inputs from either latent space; (ii) \emph{interface calibration}: train only the input and output projection layers with the transformer backbone frozen; (iii) \emph{backbone adaptation}: lightly fine-tune the transformer on 1K editing data; steps (ii) and (iii) are both trained with the flow matching loss~\citep{song2021flowmatching}, while the distillation loss in \Cref{eq:align_loss} applies only to step (i). The three-step cycle is repeated at $f{=}32$ and $f{=}64$, with the new projection warm-started from the previous factor. The entire Phase 1 completes in $\sim$4 days on 32 Kunlunxin P800 XPUs with $\sim$200K samples; the result is an initial 4K model with usable but limited quality, which Phase 2 then enhances. \Cref{fig:phase1_vs_final} compares Phase 1 output against the final model, illustrating the quality gap that motivates the enhancement phase.

\subsubsection{Phase 2: Enhancement --- Quality and Capability Refinement}
\label{subsubsection:enhancement}

\paragraph{Progressive quality SFT.}
With the initial 4K model from Phase 1, we now train the model further on high-quality editing data. Although Phase 1 aligns the representations as closely as possible, the adaptation process inevitably causes some capability degradation or forgetting in the original model, so the Phase 1 checkpoint provides a good initialization but limited editing quality.

We adopt a progressive resolution schedule: training first continues at $1024\times1024$, where data is abundant and training is efficient; then transitions to native $4096\times4096$ for the main fine-tuning phase, which is necessary because the high compression ratio causes a quality ceiling at 1K (e.g.\ persistently degraded face rendering) that only native 4K fine-tuning can break through. At each resolution, the input image is resized to a target area while preserving aspect ratio and aligned to a multiple of the VAE scale factor. This curriculum enables the DiT to gradually recover and exceed its pre-adaptation quality. During the early stage of training, we keep the data composition across task types approximately balanced to ensure broad coverage. Once the model approaches convergence, we begin dynamically adjusting the data ratio based on model performance: after each training epoch, the model is evaluated on an internal benchmark, failure modes are identified, and the data composition is rebalanced to target them. Finally, we merge checkpoints from multiple epochs via model merging to obtain a single stable model that is more robust to data distribution shifts than any individual checkpoint. This merged model is the base for downstream specialization.

\paragraph{Continual learning with ACE-LoRA and MOPD.}
Real-world editing involves a long tail of specialized needs. These needs share two characteristics: task-specific data is scarce and hard to collect, and the needs themselves are numerous. Training on such data jointly with the main SFT stage risks data dilution, which makes the specialized skills hard to learn well; learning each need separately with plain LoRA~\citep{hu2022lora} adaptation, in turn, invites task interference that drives the optimization toward inferior solutions and catastrophic forgetting that erodes the base model's general capabilities. We therefore adopt continual learning with ACE-LoRA~\citep{acelora2026}, which applies the gradient-orthogonal principle~\citep{liu2026octopus} to image editing, followed by multi-teacher on-policy distillation (MOPD)~\citep{diffusionopd2026} to consolidate the acquired knowledge back into the base model; we refer to this pair as ACE-MOPD, which makes long-sequence learning practical.

Specifically, the vertical skills are defined as many fine-grained templates, each template constituting one editing task; the templates are organized into groups of seven to nine, and each group is learned with ACE-LoRA as a two-stage gradient-orthogonal LoRA procedure: stage~1 fine-tunes the model on the group's templates, and stage~2 continues training with gradients projected onto the orthogonal complement of the subspace spanned by earlier groups' LoRAs, suppressing interference with previously learned groups. The learned LoRAs of each group then serve as teachers for one round of MOPD, which transfers the group's skills back into the base model before training moves on to the next group; the base model therefore accumulates skills group by group while the adapter count stays bounded. We assess final-task performance on held-out suites rather than assuming that the orthogonality constraint alone prevents forgetting. In principle, the adaptation--consolidation cycle can run for an unbounded number of rounds, continuously absorbing new task groups into the base model.

\begin{table}[t]
    \centering
    \small
    \caption{The 14 vertical editing tasks used for ACE-LoRA adaptation, spanning portrait beauty and three capability categories. Teeth retouching and clothing-wrinkle removal have no held-out test set.}
    \label{tab:vertical_tasks}
    \begin{tabular}{clp{5.5cm}cc}
    \toprule
    Group & Task & Sub-templates & \#Templates & \#Samples \\
    \midrule
    \multirow{9}{*}{\shortstack[l]{Portrait\\Beauty}}
        & Change Hairstyle & side-part, braids, bangs, curly, \ldots (see \Cref{tab:hairstyle_templates}) & 26 & 420 \\
        & Adjust Expression & wink, cool, closed mouth, laugh, smile, smirk, open eyes & 7 & 350 \\
        & Retouch Teeth & remove braces, remove stains, align, whiten & 4 & 305 \\
        & Adjust Gesture / Pose & stretch, heart-hands, cheek-poke, face-hold, pockets & 5 & 250 \\
        & Apply Lens Defect & reflection, smudge & 2 & 168 \\
        & Adjust Hair Volume & hairline, density boost & 2 & 84 \\
        & Apply Photo Effect & DV cam, color noise, vignette, dispersion, grain, fisheye & 6 & 300 \\
        & Apply Blur Effect & radial blur, radial burst & 2 & 100 \\
        & Remove Clothing Wrinkle & wrinkle removal & 1 & 191 \\
    \midrule
    \multirow{2}{*}{\shortstack[l]{Physical\\Perception}}
        & Refocus & --- & --- & 300 \\
        & Adjust Lighting Direction & --- & --- & 300 \\
    \midrule
    \multirow{2}{*}{\shortstack[l]{Entity\\Manipulation}}
        & Edit Text & --- & --- & 259 \\
        & Virtual Try-on & --- & --- & 294 \\
    \midrule
    \shortstack[l]{Cognitive\\Reasoning}
        & Edit Causal Relation & --- & --- & 239 \\
    \bottomrule
    \end{tabular}
\end{table}

\paragraph{RL for improving aesthetics and instruction fidelity.}
We further refine the model with reinforcement learning in two stages, targeting aesthetics and editing fidelity respectively. Both stages build on DiffusionNFT~\citep{zheng2025diffusionnft}, a likelihood-free policy optimization method that operates directly on the flow matching forward process. In the first stage, we apply DiffusionNFT with PickScore~\citep{kirstain2023pickscore}, a text-to-image preference reward, on text-to-image (T2I) tasks: aesthetics improve, but editing capability degrades, because the T2I reward favors visual appeal without accounting for instruction adherence or edit locality. We therefore continue training with BluePixReward $R(\cdot)$, our four-dimensional edit quality evaluator (detailed in \Cref{subsection:reward_model}), on image-to-image (I2I) editing tasks, which improves editing performance. To improve training efficiency and effectiveness, we filter out prompts whose sampled group has low reward variance, as these provide negligible gradient signal for policy improvement.

\subsubsection{Phase 3: Acceleration --- Few-step Distillation}
\label{subsubsection:acceleration}

The consolidated 4K model requires 50 denoising steps at inference, which is impractical for production. Recent work has shown that reinforcement learning and distillation can be combined to produce few-step generators with improved quality~\citep{rdm2026,rtdmd2026}. We attempted to apply this idea to our 4K editor, but encountered a fundamental obstacle: subtle background changes that are hardly noticeable at $1024\times1024$ become highly salient at 4K, so the reward model is heavily biased by even minute pixel-level drifts in non-edited regions, which makes RL training unstable and the model unable to improve. The mismatch is sharpest for a few-step student: reinforcement learning in Phase 2 operates on the 50-step policy, where later steps can compensate a background drift, whereas a few-step student offers no such tolerance.

This observation indicates that the core challenge in 4K editing distillation is not merely matching the output distribution of the 50-step teacher, but \emph{preserving source fidelity in non-edited regions}. We therefore build the distillation loss on \emph{source-anchored target projection} (SATP)\footnote{The full derivation and training recipe will be detailed in a companion paper.}, which introduces source-image information directly into the distillation process. The key idea is to use the teacher's prediction to distinguish edited from non-edited regions, and construct a soft target that anchors the latter to the source latent:
\begin{equation}
    \mathbf{x}_{\text{target}} = \mathbf{M} \odot \hat{\mathbf{x}}_0 + (1 - \mathbf{M}) \odot \mathbf{x}_{\text{src}},
    \label{eq:tgab}
\end{equation}
where $\hat{\mathbf{x}}_0$ is the teacher's clean-image estimate, $\mathbf{x}_{\text{src}}$ is the source latent, and $\mathbf{M}$ is an edit mask derived from the per-position edit-intensity ratio $r = \lVert \hat{\mathbf{x}}_0 - \mathbf{x}_{\text{src}} \rVert^2 \,/\, \big(\lVert \mathbf{x}_{\text{src}} \rVert^2 + \epsilon\big)$, averaged over latent channels: $\mathbf{M} = 1 - \exp(-\gamma r)$ with sensitivity $\gamma = 3$, so positions with $r \approx 0$ are anchored to the source latent while strongly edited positions retain the teacher's estimate; in practice $\exp(-\gamma r)$ is binarized at $0.5$ to yield a hard keep/edit partition. The student is then fine-tuned by combining the standard distillation loss with a fidelity term on this anchored target (\Cref{eq:tgab}). By explicitly enforcing source fidelity during distillation, SATP compensates for the sensitivity that made RL infeasible.

The resulting 4-step student preserves task success rate and background consistency relative to the 50-step teacher, while maintaining comparable editing quality; it also achieves a $\sim\!4\times$ inference speedup: with bf16 inference and no other optimization, a 4K edit completes in 8.7\,s on a single Kunlunxin P800 XPU. The distilled 4K model serves as the base editor for the agentic harness described in \Cref{subsection:harness}, where it is exposed as the primary editing skill.

%% ================================================================
%%  AGENTIC EDITING HARNESS
%% ================================================================

\subsection{Agentic Editing Harness}
\label{subsection:harness}

\paragraph{Overview.}

The 4K diffusion model from \Cref{subsection:4k_training} provides the visual substrate, but it remains a single model: one network takes an image and an instruction and returns one edited result. As editing tasks grow more compositional (multi-turn refinement, region-disjoint edits, and reference-guided stylization), a single model reaches a capability ceiling that additional training data alone cannot resolve. \bpxmodel addresses this by wrapping the 4K model inside an \emph{agentic editing harness} (\emph{BluePixel Agent}) that decomposes an editing request into a routed sequence of composable skills with closed-loop verification. The design follows two principles: (i) \emph{composability}: editing is performed by a registry of specialized, independently replaceable skills rather than a single monolithic model; and (ii) \emph{closed-loop verification}: every edit is evaluated by a quality module before being returned, and failures trigger automatic re-editing instead of silent acceptance.

\subsubsection{System Architecture}
\label{subsubsection:architecture}

BluePixel Agent organizes editing around three complementary functions:

\begin{enumerate}\itemsep0pt
    \item \textbf{Request interpretation.} A router considers the user's instruction together with the current image and conversation context. It determines whether there is enough information to proceed with an edit or whether the turn instead calls for clarification, suggestions, or assessment of an existing result. For an actionable edit, an intent analyst formulates the operations to be performed and may gather supporting information for underspecified references. For example, when a user names a visual style without describing its defining features, the analyst may use web search to identify relevant visual attributes before specifying the edit.
    \item \textbf{Skill execution.} The harness applies the relevant editing skills to these operations (\Cref{tab:skills}). Specialized tools handle constrained tasks such as color grading and template-guided editing, while the general-purpose editor handles open-ended changes to image content.
    \item \textbf{Feedback and selection.} The resulting images are assessed against the instruction and source image. This feedback drives further editing when needed and the final candidate selection (\Cref{subsubsection:closed_loop}).
\end{enumerate}

\begin{figure*}[!t]
    \centering
    \includegraphics[width=\textwidth,keepaspectratio]{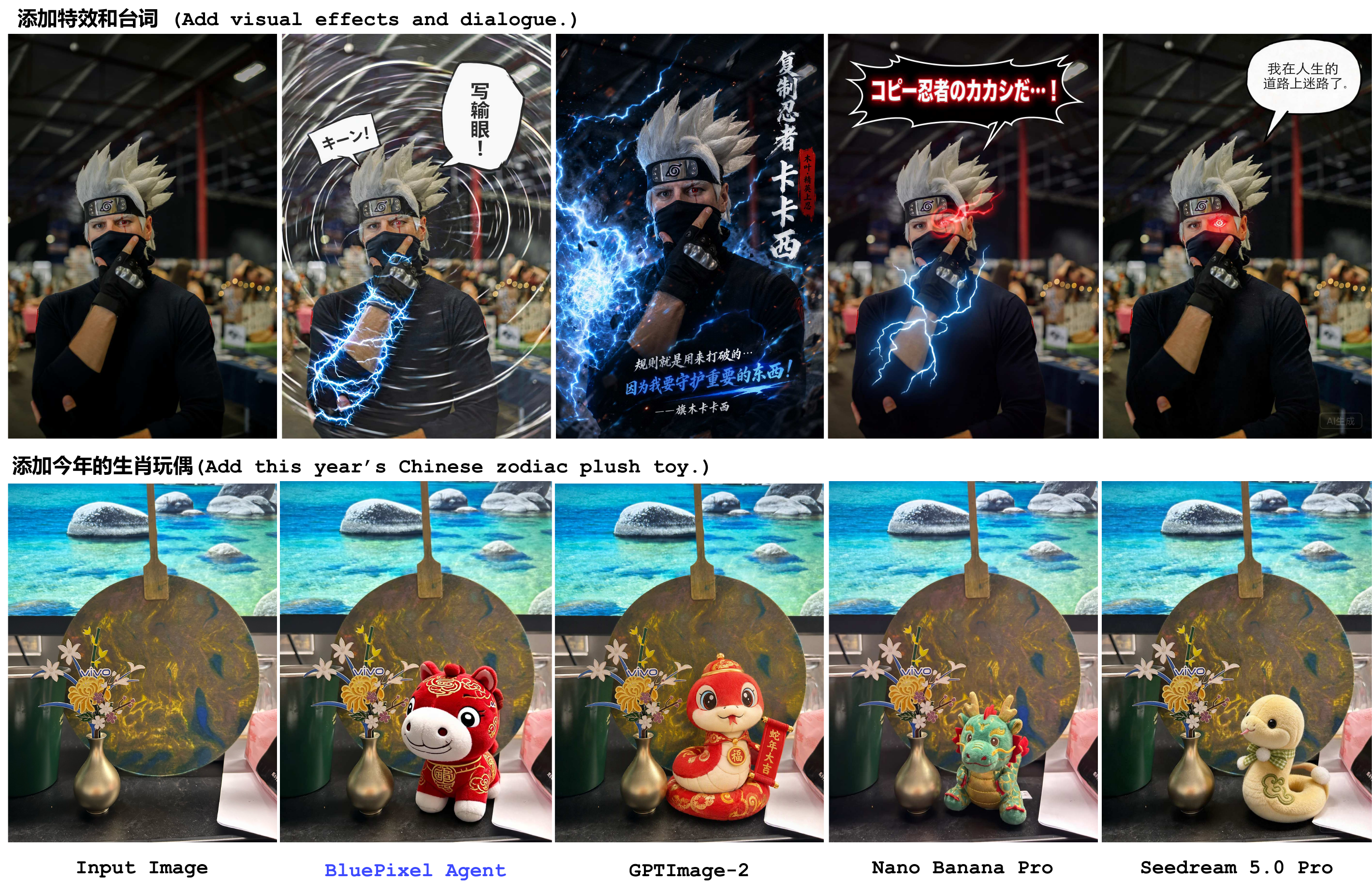}
    \caption{\textbf{Agent-assisted editing with supplementary factual context.} Top: adding visual effects and dialogue associated with the depicted character. Bottom: adding a plush toy corresponding to the Chinese zodiac year specified in the instruction. Each row shows, left to right: the input image, BluePixel Agent, GPT Image 2, Nano Banana Pro, and Seedream 5.0 Pro.}
    \label{fig:agent_grounding}
\end{figure*}

Some edit requests depend on facts beyond the source image, such as attributes of a named character or a time-sensitive reference. In these cases, the intent analyst can use web search to gather relevant information and incorporate it into the editing instruction. \Cref{fig:agent_grounding} illustrates this capability with character-specific visual effects and dialogue, and with a zodiac-themed toy requested by reference to the current year.

These functions separate decisions about what should be edited, how the edit is carried out, and how its outcome is assessed. The harness also retains conversational constraints and image provenance across turns, allowing users to refine a selected result or branch from an earlier version. This continuity connects heterogeneous editing skills within a service rather than treating each request as an isolated model call.

\begin{figure}[!t]
    \centering
    \includegraphics[width=\linewidth]{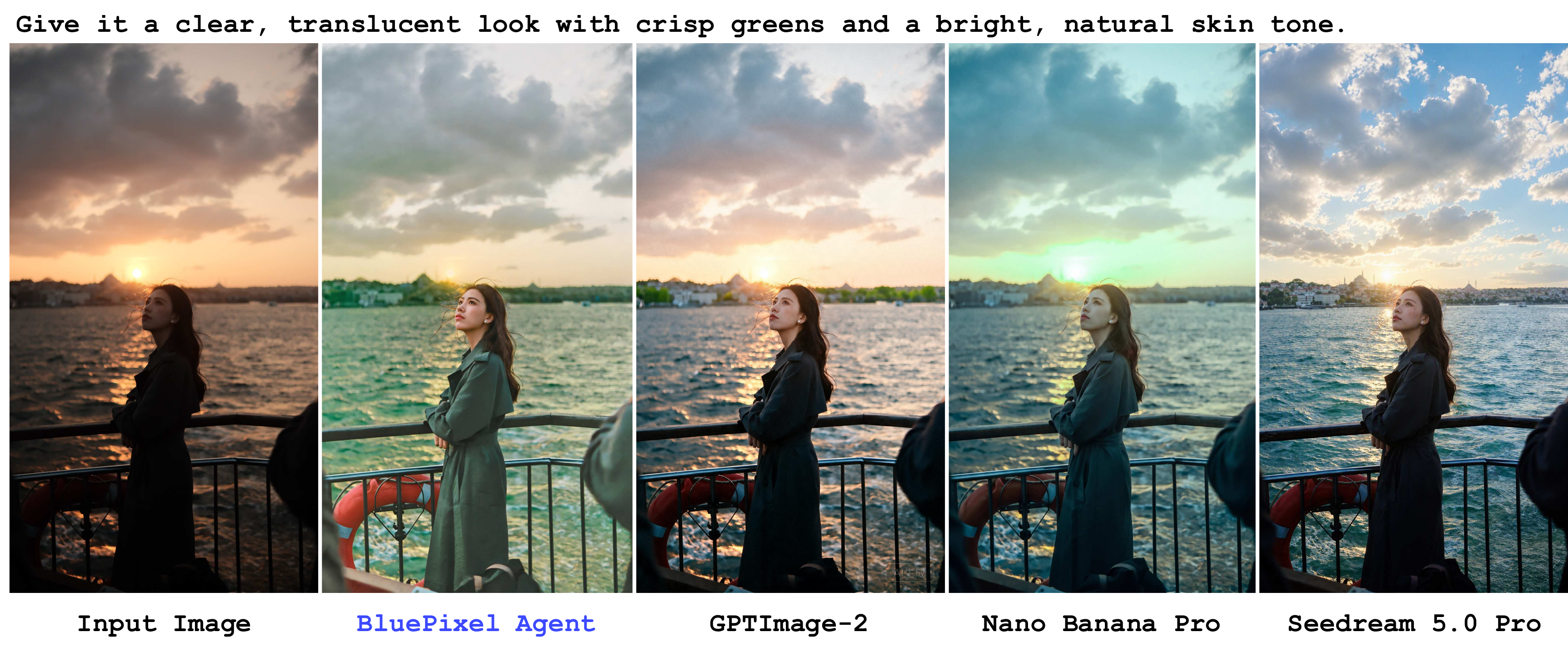}
    \caption{\textbf{Filter-style editing comparison.} BluePixel Agent vs.\ GPT Image 2, Nano Banana Pro, and Seedream 5.0 Pro on a filter task requiring professional aesthetic judgment. The three compared baselines produce plausible but aesthetically inconsistent results. BluePixel Agent routes the request to LUT-based (look-up-table) color grading, a specialist editing capability in \Cref{tab:skills}, rather than relying solely on the diffusion model; this produces more accurate and professionally consistent filter effects.}
    \label{fig:filter_comparison}
\end{figure}

The advantage of the skill-based approach is illustrated in \Cref{fig:filter_comparison}. When a user requests a specific filter look (a task with strong professional aesthetic orientation), a pure diffusion model tends to generate a plausible but imprecise approximation. BluePixel Agent, by contrast, can use LUT-based color grading to apply a calibrated color transform that matches the intended look. This is a direct consequence of the composable skill design: tasks that have well-defined technical solutions are handled by dedicated tools, while the primary editing model is reserved for generative edits that truly require it. We next describe the skill-based editing repertoire and the feedback process used to assess its outputs.

\subsubsection{Skill-Based Editing}
\label{subsubsection:skill_registry}

Production-grade image editing spans a wider range of operations than a single image-to-image transformation. A request may call for global stylization, localized changes, a calibrated color treatment, or a template-guided transformation; these operations place different demands on how an image is modified. BluePixel Agent therefore treats the 4K diffusion editor as one capability within a broader skill repertoire rather than the sole mechanism for every edit.

The repertoire has three functional tiers, as shown in \Cref{tab:skills}. The \emph{primary editing} tier contains the BluePixel 4K diffusion editor for open-ended changes to image content. \emph{Specialist editing skills} encapsulate operations tailored to particular editing domains, with task-specific controls that can be used independently when a request falls within their scope. When an instruction combines such an operation with open-ended content changes, the harness can coordinate the specialist skill with the primary editor. \emph{Auxiliary capabilities} interpret intent, resolve references, refine instructions, and assess results; they guide the editing process without directly modifying the image.

\begin{table*}[t]
    \centering
    \small
    \caption{\textbf{Skill-based editing capabilities.} Representative editing and supporting capabilities of BluePixel Agent, organized by functional tier.}
    \label{tab:skills}
    \begin{tabular}{@{}>{\raggedright\arraybackslash}p{1.6cm}l>{\raggedright\arraybackslash}p{8.4cm}@{}}
        \toprule
        \textbf{Tier} & \textbf{Capability} & \textbf{Role} \\
        \midrule
        Primary editing & BluePixel 4K diffusion editor & Native-4K, open-ended edits to objects, backgrounds, and scene composition \\
        \midrule
        \multirow{6}{=}{\raggedright Specialist editing} & LUT-based color grading & Global color transformations for filter-style edits \\
        & Tone and lighting effects & Preset-based adjustments to illumination and visual atmosphere \\
        & Template-guided editing & Application of predefined styles or scene treatments \\
        & Mask-guided local editing & Changes restricted to a selected image region \\
        & Hairstyle editing & Hair-region edits guided by a user-provided mask \\
        & Reference-guided prop editing & Placement of a reference object within a designated region \\
        \midrule
        \multirow{4}{=}{\raggedright Auxiliary} & Intent and reference grounding & Interprets the request and resolves underspecified references \\
        & In-house prompt refinement model & Reformulates instructions before editing or using evaluation feedback \\
        & BluePixReward evaluator & Assesses instruction following, content preservation, and visual quality \\
        & Version tracking & Retains image history for multi-turn refinement \\
        \bottomrule
    \end{tabular}
\end{table*}

\subsubsection{Closed-Loop Editing}
\label{subsubsection:closed_loop}

Editing in BluePixel Agent is not an open-loop chain that emits a single result and trusts it; it is a \emph{closed-loop} process: edit, verify, and re-edit if necessary. An edited image is treated as a candidate whose assessment informs both subsequent actions and the final result.

Once the requested edit and its constraints have been specified, the \emph{image editor} executes the relevant skill and produces a candidate. The \emph{quality evaluator} receives the source image, candidate, and editing instruction, and uses BluePixReward, our VLM-based judge (\Cref{subsection:reward_model}), to assess instruction following, non-edited region consistency, and visual quality. It returns scores and diagnostic feedback that the harness can use to decide what to do next.

The evaluation is actionable rather than merely descriptive. If a specialist edit leaves part of the request unmet, the harness can retry the same skill and, if needed, carry the unmet requirement into a general edit. For a generative edit, the diagnostic feedback can inform a revised instruction for another attempt. Recovery attempts are bounded, and the harness retains intermediate candidates rather than assuming that the latest edit is the best. It selects the final image by assessing the candidates against the complete request, so feedback governs both whether editing continues and which result is returned.

%% ================================================================
%%  SECTION 5: DATA CONSTRUCTION AND QUALITY ASSESSMENT
%% ================================================================

\section{Data Construction and Quality Assessment}
\label{section:data_pipeline}

\subsection{Overview}

\begin{figure}[!t]
    \centering
    \includegraphics[width=\linewidth]{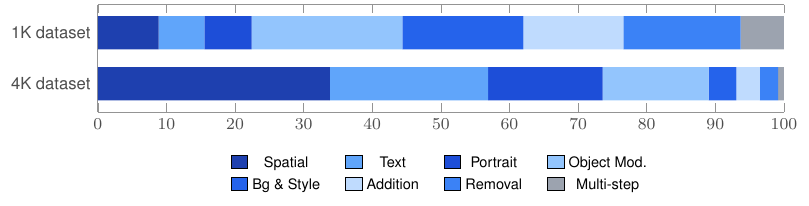}
    \caption{\textbf{The 4K set concentrates on three task types}, while the 1K set is more evenly distributed: task type proportions for the 1K and 4K training datasets across the eight unified task types.}
    \label{fig:dataset_ratio}
\end{figure}

Our training data combines publicly available editing datasets, including UnicEdit-10M~\citep{ye2025unicedit}, TextEdit~\citep{tian2026internvlu}, X2Edit~\citep{ma2025x2edit}, and SpatialEdit~\citep{xiao2026spatialedit}, with in-house annotated pairs. Because Phase 1 yields a good initialization, we adopt a quality-over-quantity strategy: rather than scaling data volume, we focus on filtering for visual quality, aesthetics, information density, and edit accuracy. The pipeline applies two quality gates: an \emph{image quality filter} that discards pairs failing visual quality checks, and an \emph{edit evaluation reward model} (BluePixReward, detailed in \Cref{subsection:reward_model}) that scores each surviving triple (source, edited, instruction) on instruction following, non-edited region consistency, visual quality, and overall. The same quality standard governs both training-time filtering and evaluation. We construct 1K and 4K datasets, reorganized into eight task types: Spatial Editing, Text Editing, Portrait Editing, Object Modification, Background \& Style Editing, Object Addition, Object Removal, and Multi-step Editing (\Cref{fig:dataset_ratio}).

\subsection{Image Quality Filtering}

VINS-120K~\citep{guan2026vins} establishes a multi-stage filtering paradigm for high-resolution editing data: preliminary file checks, image quality filtering via hand-crafted perceptual operators (sharpness, exposure, color, and texture statistics), instruction-following filtering via VLM parsing combined with object detection and segmentation, and aesthetic assessment via dedicated predictors; each stage relies on a manually tuned threshold, and in VINS-120K only the top 20\% of samples survive the full pipeline. We adopt the same joint filtering principle (if either the source or the edited image fails any quality check, the pair is discarded), but observe that per-operator threshold tuning is inherently subjective, does not generalize across data distributions, and requires maintaining a large stack of independent operators with their own failure modes.

To eliminate this subjectivity, we train a single VLM based on Qwen3-VL-8B~\citep{bai2025qwen3vl} that takes the image as input and directly outputs scores for multiple quality metrics along with a reasoned analysis, producing its own retain-or-discard judgment. The model evaluates five axes that capture complementary aspects of image quality: \emph{aesthetics} (composition, lighting, and color harmony), \emph{visual clarity} (image sharpness and artifacts), \emph{saturation} (color vividness and richness), \emph{information density} (texture, structural details, and semantic complexity), and \emph{watermark presence}. We then filter images based solely on the model's judgment, replacing the brittle stack of hand-tuned thresholds with a single, learned quality gate. We pay particular attention to information density: high-density data, comprising images rich in texture, structure, and semantic content, can accelerate convergence, as each sample exposes the model to more diverse visual patterns.

\subsection{Edit Evaluation Reward Model}
\label{subsection:reward_model}

Beyond general image quality, we need to verify that each edit pair is \emph{correct}: the edited image must faithfully follow the instruction without introducing artifacts or damaging non-edited regions. To this end, we train a dedicated edit evaluation reward model, \textbf{BluePixReward}. Expert models are first trained on different editing tasks, and multi-teacher on-policy distillation (MOPD)~\citep{diffusionopd2026} then distills them into a single judge built on Qwen3.5-9B; it takes the source image, edited image, and instruction as a triple and outputs four-dimensional scores\footnote{The training recipe for the reward model will be detailed in a companion paper.}:
\begin{itemize}\itemsep0pt
    \item \textbf{Instruction Following (IF):} whether the edit accurately executes the instruction, penalizing wrong edit type, missing or incomplete edits, extraneous modifications, and person-related errors.
    \item \textbf{Non-Edited Region Consistency (NC):} requires a reasoning step that first localizes the edit region, then compares the complement; camera-motion edits are treated as a special case.
    \item \textbf{Visual Quality (VQ):} naturalness and fusion quality of the edited region and overall image, penalizing fusion seams, material/lighting mismatches, inpainting artifacts, person defects, and severe distortion across the whole image.
    \item \textbf{Overall:} a composite quality score reported alongside the three component scores. The evaluator also provides a binary success decision: an output counts as successful when $\mathrm{IF}\ge 9$, $\mathrm{NC}\ge 8$, and $\mathrm{VQ}\ge 6$, based on its assessment of instruction following, non-edited-region consistency, and visual quality. This decision is reported as the Success column in \Cref{tab:vins}.
\end{itemize}
The scores are calibrated against human judgment: Pearson correlations with expert Likert ratings on a 150-pair validation set are $r=0.82$ (IF), $r=0.78$ (NC), $r=0.76$ (VQ), and $r=0.84$ (Overall). This reward model serves as the \emph{unified quality signal} throughout the pipeline: low-scoring pairs are discarded from training data; multiple candidates for the same instruction are scored jointly as a group, providing the reward signal for DiffusionNFT; the same model also serves as the benchmark evaluator for VINS-4KEval++ (\Cref{tab:vins}, the BluePixReward protocol), whose four-dimensional scores serve as the primary evaluation metrics, closing the quality loop.

\subsection{4K Data Construction}

Similar to VINS-120K~\citep{guan2026vins}, we source 4K pairs from two complementary streams, with the majority coming from video mining. The first uses a dedicated pipeline that mines video footage for frame pairs that can serve as editing data, selecting frames where the visual difference between two shots constitutes a plausible edit, and generating corresponding instructions via prompt rewriting. The second applies super-resolution to a subset of 1K source images. Before upscaling, each candidate is screened by MANIQA~\citep{maniqa}, a no-reference image-quality model used here for sharpness screening: the image is cropped into $224\times224$ patches, each patch is scored with a Vision Transformer (ViT), and the patch scores are aggregated; upscaling a blurry 1K image produces artifact-laden 4K results, so only sufficiently sharp images proceed. All 4K pairs pass through the same image quality filtering and edit evaluation reward model as their 1K counterparts.

%% ================================================================
%%  SECTION 6: EXPERIMENTS
%% ================================================================

\section{Experiments}
\label{section:experiments}

\paragraph{Benchmarks.}
We evaluate BluePixel on four benchmarks. \textbf{ImgEdit-Bench}~\citep{ye2025imgedit} provides broad task coverage across nine editing categories at standard resolutions; for fair comparison, we resize our results to $1024\times1024$ on this benchmark. Next, \textbf{VINS-4KEval}~\citep{guan2026vins} is the first benchmark dedicated to 4K image editing, and we use it to evaluate the performance at native 4K resolution. Third, we extend VINS-4KEval with more diverse scenes and instructions, termed \textbf{VINS-4KEval++}, comprising 3{,}305 evaluation cases. Finally, we use \textbf{CIE-Bench}~\citep{acelora2026} to evaluate the continual learning performance (as introduced in \Cref{subsubsection:enhancement}).

\subsection{Main Results}
\label{subsection:main_results}

\subsubsection{Qualitative Results}
\label{subsubsection:qualitative}

\begin{figure*}[!t]
    \centering
    \includegraphics[width=\textwidth]{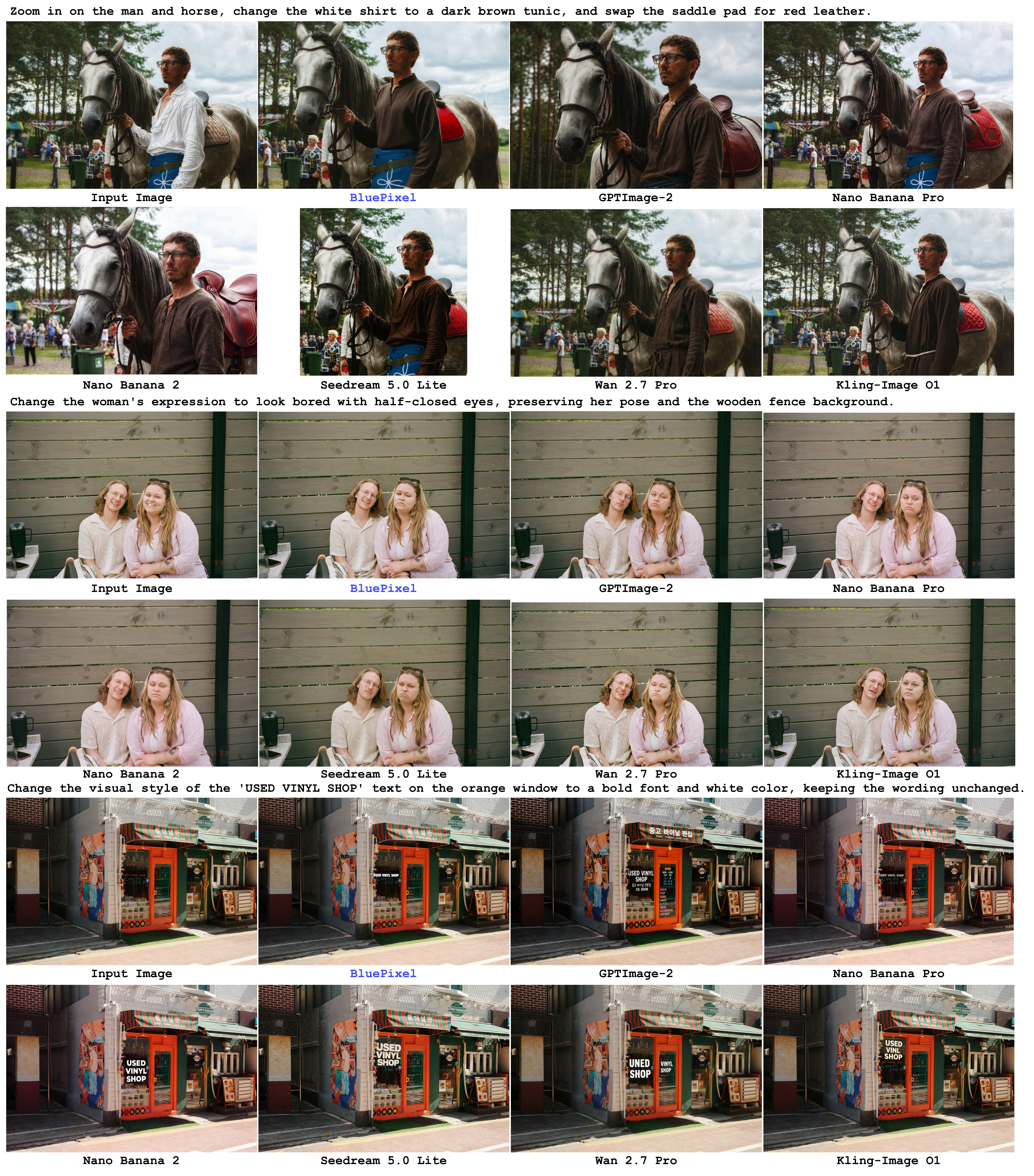}
    \caption{\textbf{Main qualitative results on VINS-4KEval++.} Each two-row block shows one instruction: compound attribute editing, facial-expression editing, or text restyling. Columns, left to right: the input image, BluePixel, GPT Image 2, Nano Banana Pro, Nano Banana 2.0, Seedream 5.0 Lite, Wan 2.7 Pro, and Kling.}
    \label{fig:qualitative}
\end{figure*}

\paragraph{Comparison on VINS-4KEval++.}
\Cref{fig:qualitative} compares BluePixel with six representative baselines (GPT Image 2~\citep{gptimage}, Nano Banana Pro~\citep{nanobanana}, Nano Banana 2.0~\citep{nanobanana}, Seedream 5.0 Lite~\citep{seedream5}, Wan 2.7 Pro~\citep{wan27pro}, and Kling~\citep{klingai}) on three examples drawn from VINS-4KEval++ at native 4K. The instructions require compound attribute edits, facial-expression manipulation, and typographic restyling without changing the text content. In the attribute-editing example, all six baselines render the dark brown tunic, but GPT Image 2 alters the fairground background and restyles the whole saddle instead of only making the pad red leather, whereas BluePixel makes exactly the requested changes and preserves the rest of the scene. In the expression example, the strongest baselines reach a similar bored look, but GPT Image 2 also alters the man beside her (eyes closed, smile gone) although the instruction targets only the woman; BluePixel edits her alone. The text-editing example shows the widest gap: Wan 2.7 Pro retypes the sign as ``UNED SHOP'' and Kling misspells it as ``USED VINL SHOP'', whereas BluePixel renders the requested bold white lettering with the wording and the storefront layout untouched, sharp at full resolution.

\begin{figure*}[!t]
    \centering
    \includegraphics[width=\textwidth]{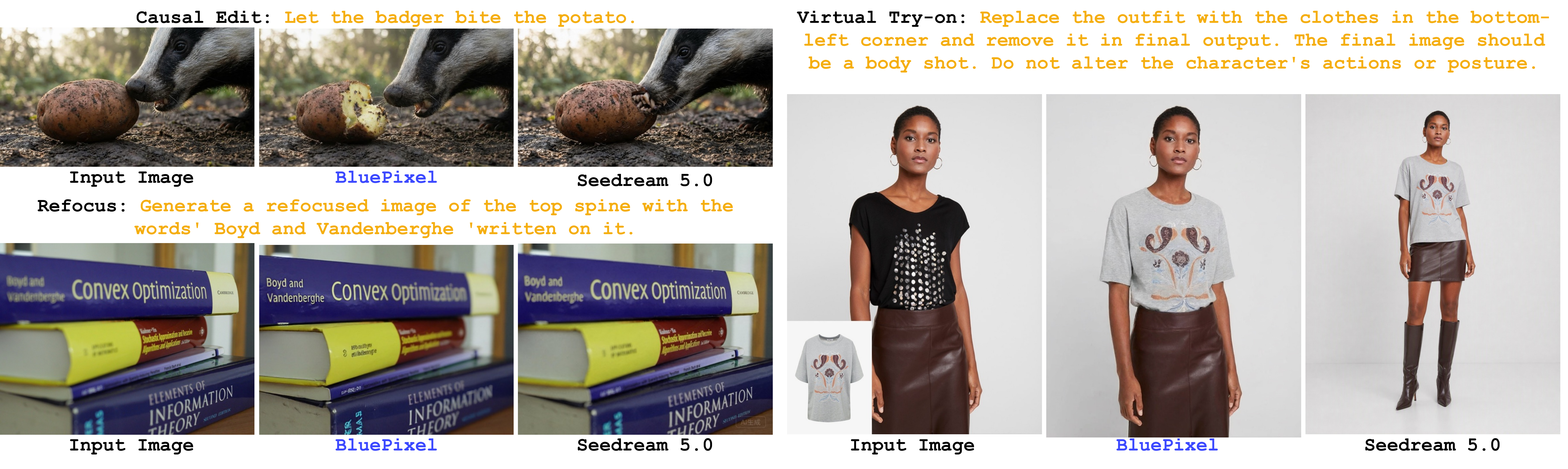}
    \caption{\textbf{Qualitative comparison on CIE-Bench~\citep{acelora2026}.} BluePixel vs.\ Seedream 5.0 on a causal-relation edit and a refocus edit (left), and a virtual try-on (right). BluePixel renders each requested outcome --- the badger's bite, the legible refocused spine text, and the garment transfer at the original framing --- whereas Seedream 5.0 shows an intact potato, unchanged spine text, and a full-body shot.}
    \label{fig:ciebench}
\end{figure*}

\paragraph{Effect of vertical-template learning.}
On the specialized-skill side, \Cref{fig:ciebench} compares the BluePixel model with Seedream 5.0~\citep{seedream5} on three vertical tasks: a causal-relation edit, a refocus edit, and a virtual try-on. BluePixel realizes the requested causal outcomes (a bite taken out of the potato, and the requested words rendered legibly on the refocused spine) and transfers the reference garment while preserving the original framing and posture, whereas Seedream 5.0 renders the badger in contact with an intact potato, leaves the spine text at its original size, and reframes the try-on result as a full-body shot. These tasks are precisely the vertical skills added by continual learning, which we analyze next.

\subsubsection{Model Analysis}
\label{subsubsection:model_analysis}

Beyond the end-to-end comparison, we analyze the two components that distinguish BluePixel from a plain editing model: the agent harness at inference and continual learning with ACE-LoRA and MOPD in training.

\begin{figure*}[!t]
    \centering
    \includegraphics[width=\textwidth]{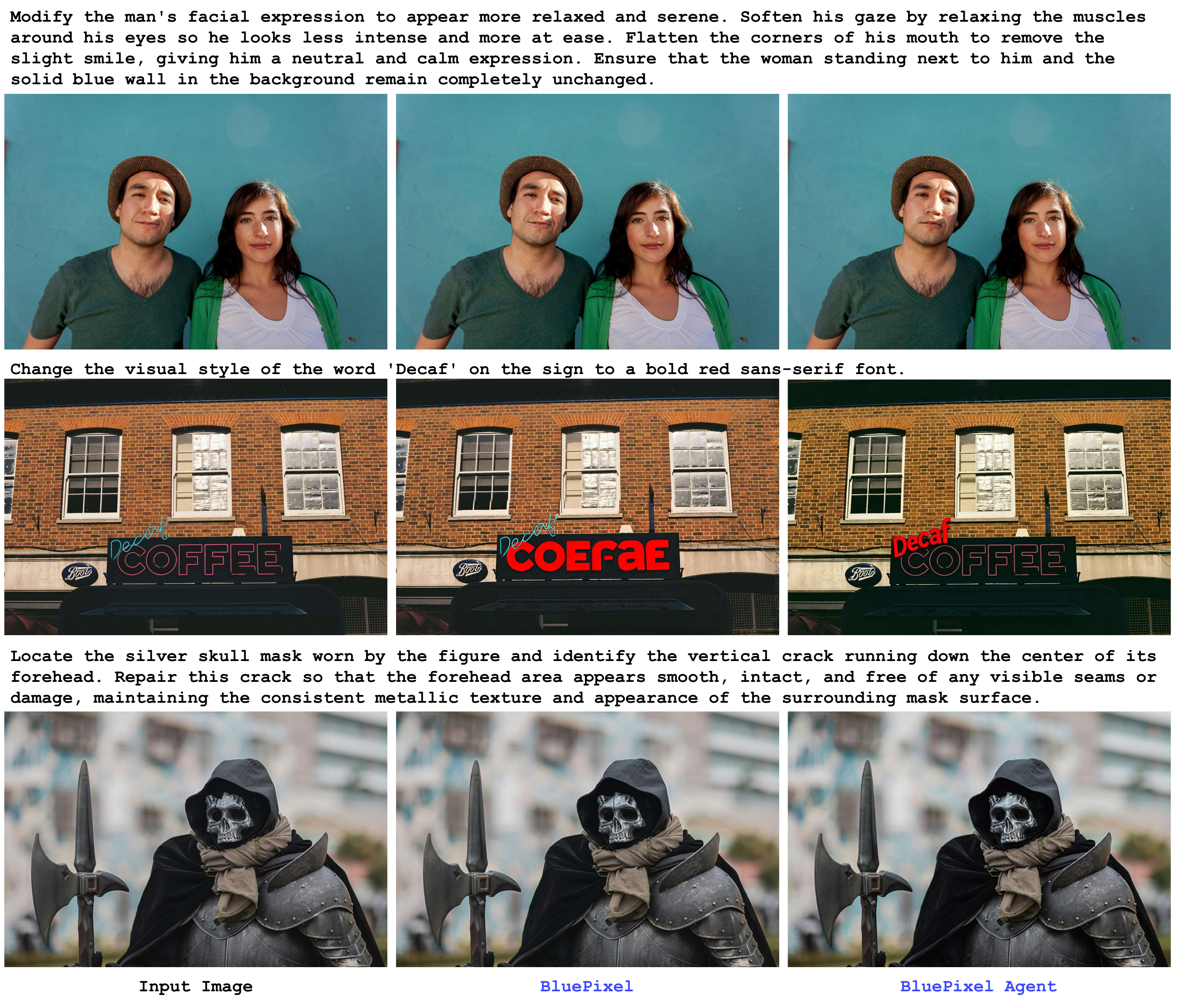}
    \caption{\textbf{Effect of the agent harness.} Three editing instructions (facial expression, text restyling, and mask repair) executed by the BluePixel backbone alone (middle) and by BluePixel Agent (right). The agent produces the requested expression, renders the target word correctly, and fully repairs the cracked mask, while the backbone under-modifies or corrupts these details.}
    \label{fig:agent_comparison}
\end{figure*}

\paragraph{Is the agent harness effective?}
\Cref{fig:agent_comparison} isolates the contribution of the agent harness by contrasting the backbone editor with BluePixel Agent on the same instructions. In the facial-expression example, the backbone changes the expression in the right direction but overshoots, producing a stiff, grinning look rather than the requested relaxed and serene one; the agent's verification loop catches the mismatch, retries, and returns a calm expression that matches the instruction. In the text example, the backbone ignores half of the request: the word is rendered as garbled letterforms (``COEFAE'') in the wrong color and font, whereas the agent produces ``Decaf'' in bold red sans-serif without disturbing the surrounding sign. In the mask-repair example, the backbone leaves the crack clearly visible across the mask's forehead, whereas the agent repairs the fracture and restores a continuous metallic texture. Across all three cases the backbone delivers an approximate one-shot result, and the agent converges to an output that actually satisfies the instruction, which is precisely what the verification-and-retry loop is for.

\begin{figure*}[!t]
    \centering
    \includegraphics[width=\textwidth]{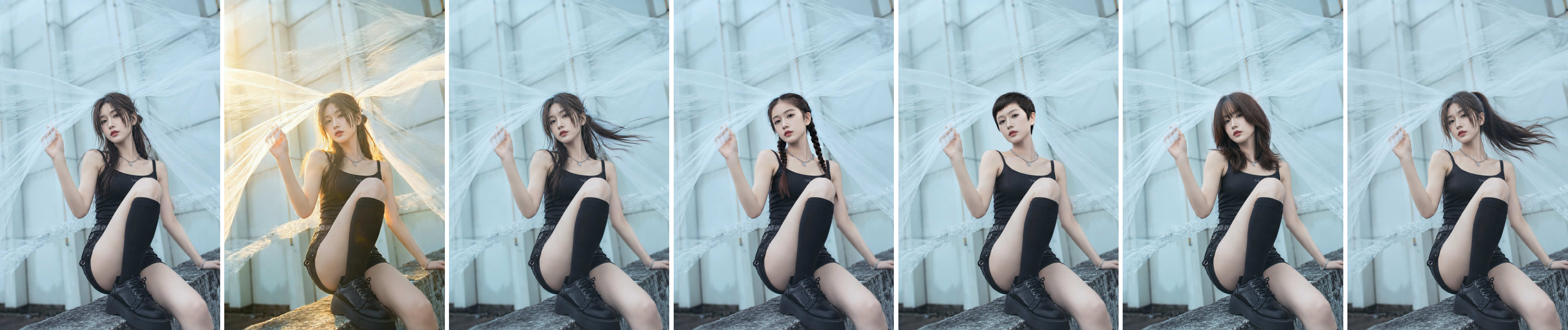}
    \caption{\textbf{ACE-MOPD qualitative visualization: hairstyle manipulation.} The leftmost column shows the source image; subsequent columns correspond to six target hairstyles: wind-swept long hair under warm backlight, wind-swept long hair, braids, a short pixie cut, loose waves, and a high ponytail. The edits change hair structure and appearance while retaining subject identity, apparel, pose, and scene layout.}
    \label{fig:vertical-hair}
\end{figure*}

\begin{figure*}[!t]
    \centering
    \includegraphics[width=\textwidth]{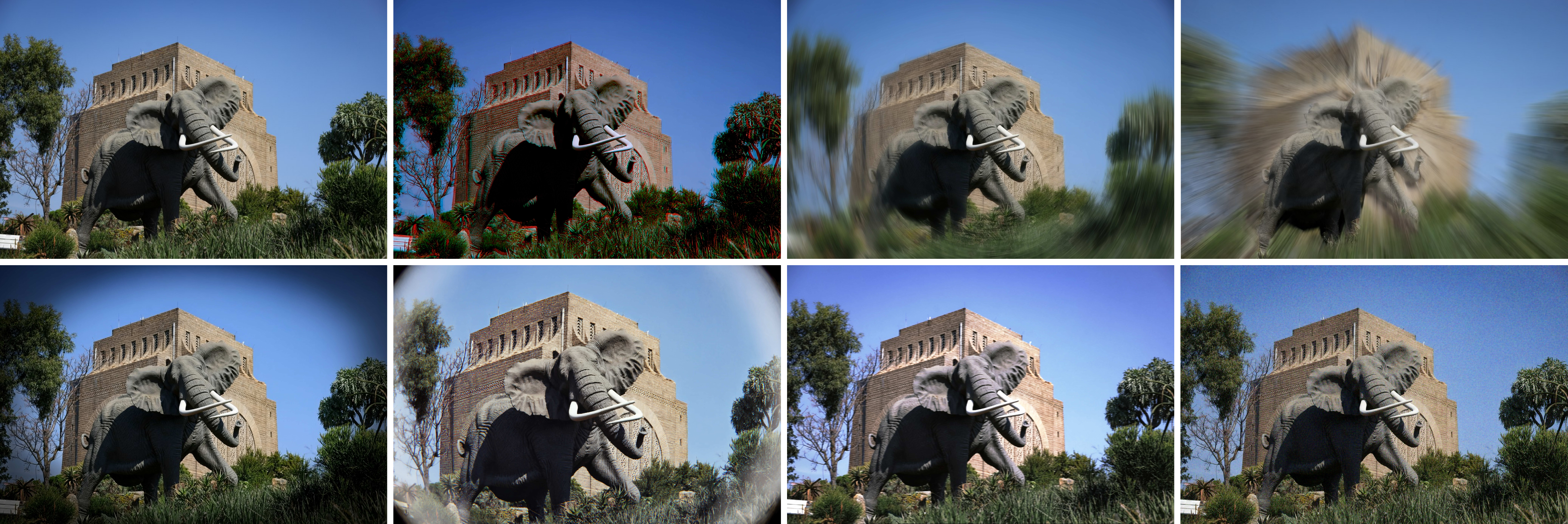}
    \caption{\textbf{ACE-MOPD qualitative visualization: global image-effect rendering.} The leftmost column shows the source image; subsequent columns apply chromatic aberration, directional motion blur, radial zoom blur, vignetting, fisheye distortion, a DV-camera effect, and chromatic noise, respectively. The source image's semantic content and scene structure are retained under these global appearance transformations.}
    \label{fig:vertical-effects}
\end{figure*}

\begin{figure*}[!t]
    \centering
    \includegraphics[width=\textwidth]{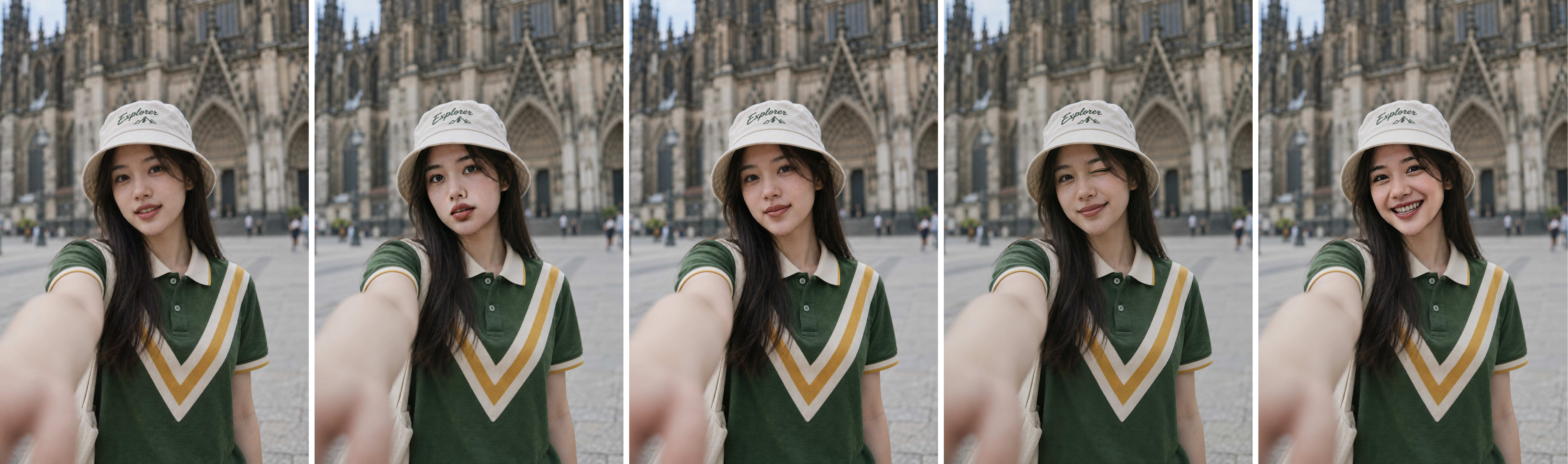}
    \caption{\textbf{ACE-MOPD qualitative visualization: facial-expression manipulation.} The leftmost column shows the source image; subsequent columns depict a surprised expression, a subtle smile, a wink, and a broad smile. The transformation is localized to the face while preserving identity cues, hairstyle, apparel, viewpoint, and background.}
    \label{fig:vertical-expression}
\end{figure*}

\begin{figure*}[!t]
    \centering
    \includegraphics[width=\textwidth]{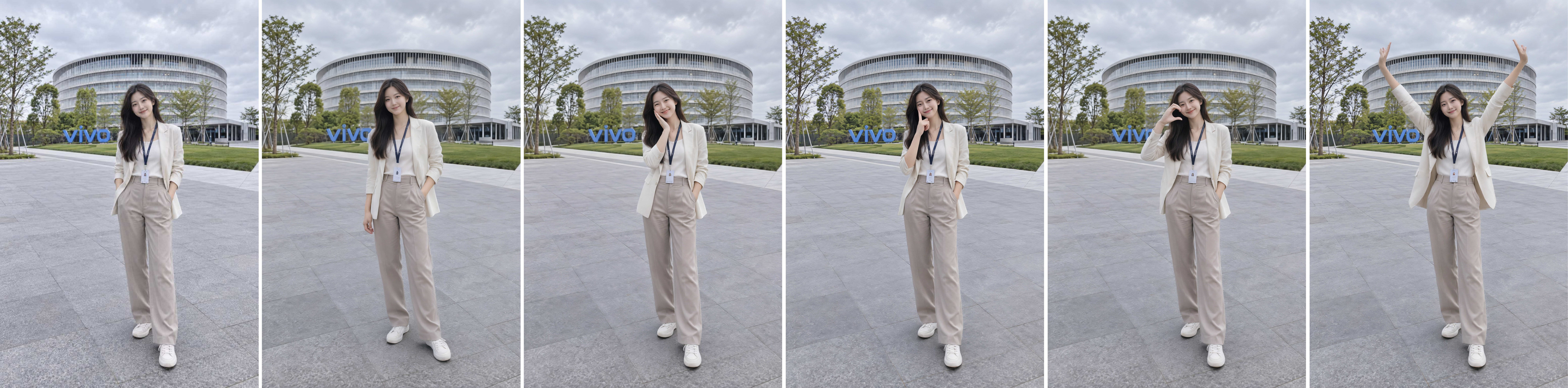}
    \caption{\textbf{ACE-MOPD qualitative visualization: human pose manipulation.} The leftmost column shows the source image; subsequent columns show a relaxed standing pose, a cheek-heart gesture, a head tilt, a peace sign, and a raised-arms pose. The model changes the articulated body configuration while maintaining subject identity, clothing, full-body framing, and the outdoor scene.}
    \label{fig:vertical-pose}
\end{figure*}

\paragraph{Can ACE-MOPD absorb data-scarce yet practically valuable templates?}
The fine-grained templates of \Cref{tab:vertical_tasks} are exactly that: data-scarce yet practically valuable. Each is learned from only about 15 to 200 samples, yet covers recurring creative-workflow needs (hairstyle change, expression control, virtual try-on, causal-relation editing), where the instruction specifies the target attribute or effect and everything outside it (identity, apparel, scene content, viewpoint) is a preservation constraint. \Cref{fig:vertical-hair,fig:vertical-effects,fig:vertical-expression,fig:vertical-pose} present 22 representative outputs of the ACE-MOPD continual-learning model, spanning local semantic control (hairstyle and expression) and broader geometric or photometric transformations (pose and global effects). They show that BluePixel learns a large repertoire of such templates and retains strong few-shot learning capability over a long sequence of tasks.

\begin{figure}[h]
\centering
\begin{minipage}[b]{0.52\linewidth}
\centering
{\footnotesize\setlength{\tabcolsep}{2.5pt}
\begin{tabular}{lccc}
\toprule
Task & SFT$\uparrow$ & $+$ ACE-MOPD$\uparrow$ & $\Delta$ \\
\midrule
Refocus & 7.53 & 8.71 & +1.19 \\
Light & 3.19 & 7.87 & +4.69 \\
Text editing & 8.37 & 9.28 & +0.91 \\
Causal-relation editing & 7.50 & 9.42 & +1.92 \\
Virtual try-on & 4.60 & 9.05 & +4.45 \\
\midrule
Average & 6.24 & 8.87 & +2.63 \\
\bottomrule
\end{tabular}}
\captionof{table}{CIE-Bench capability tasks before and after continual learning (benchmark-native VLM-judge protocol; scores are the equal-weight average over the five tasks).}
\label{tab:cie_pertask}
\end{minipage}\hfill
\begin{minipage}[b]{0.44\linewidth}
\centering
\includegraphics[width=0.9\linewidth]{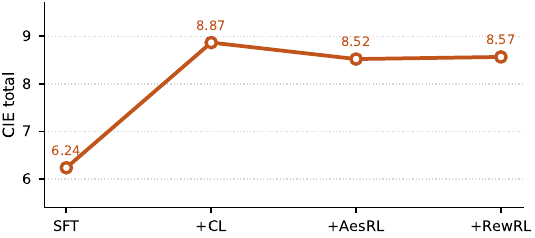}
\captionof{figure}{Stage-wise CIE total across the training recipe (SFT, $+$ ACE-MOPD, $+$ aesthetic RL, $+$ reward RL; single run per stage, benchmark VLM judge).}
\label{fig:cie_traj}
\end{minipage}
\end{figure}

On the five CIE-Bench capability tasks, scored under the benchmark's native VLM-judge protocol (instruction adherence $+$ naturalness, 10-point total), continual learning improves every capability task (\Cref{tab:cie_pertask}): the lighting task gains most (+4.69), followed by virtual try-on (+4.45), while the portrait-beauty suite shows no regression. Across the training stages, the CIE total rises from 6.24 in the base model to 8.87 after continual learning with ACE-LoRA and MOPD, dips to 8.52 after aesthetic RL, and partially recovers to 8.57 after reward RL with BluePixReward (\Cref{fig:cie_traj}); general-editing quality on VINS-4KEval++ follows the opposite trajectory across the same stages.

\paragraph{Seed stability and per-task-type breakdown.}
\Cref{tab:seed} lists the eight seeds behind the mean in \Cref{tab:vins}: the per-seed Overall spans 0.053 (noise level); no seed falls below Nano Banana 2.0 (8.561), and every seed except one matches or exceeds Nano Banana Pro (8.601), so the 0.02-point margin over Nano Banana Pro is within seed noise. \Cref{tab:tasktype} breaks the same runs down by the benchmark's nine task types: adding objects is the strongest category (Overall 9.20), changing camera viewpoint the weakest (7.55), largely because instruction following conflicts with non-edited-region preservation.

\begin{table}[h]
\centering
\small
\caption{Per-seed results on the 2{,}513 shared VINS-4KEval++ cases (BluePixReward judge, prompt rewriting), ranked by Overall. $\Delta$ is the Overall margin over Nano Banana Pro.}
\label{tab:seed}
\setlength{\tabcolsep}{4pt}
\begin{tabular}{lrrr}
\toprule
Seed & Overall$\uparrow$ & Success (\%)$\uparrow$ & $\Delta$ vs.\ Nano Banana Pro \\
\midrule
2026 & 8.653 & 63.8 & +0.052 \\
42 & 8.651 & 64.0 & +0.050 \\
1337 & 8.621 & 63.2 & +0.020 \\
1 & 8.619 & 63.4 & +0.018 \\
3407 & 8.619 & 62.4 & +0.018 \\
777 & 8.615 & 63.0 & +0.014 \\
1234 & 8.612 & 62.9 & +0.011 \\
0 & 8.600 & 62.9 & $-$0.001 \\
\midrule
8-seed mean & 8.624 & 63.2 & +0.023 \\
\bottomrule
\end{tabular}
\end{table}

\begin{table}[h]
\centering
\small
\caption{Per-task-type results on the full VINS-4KEval++ benchmark (8-seed mean, BluePixReward judge).}
\label{tab:tasktype}
\begin{tabular}{lrr}
\toprule
Task type & Overall$\uparrow$ & Success (\%)$\uparrow$ \\
\midrule
Add object or element & 9.203 & 79.3 \\
Remove object or element & 8.994 & 72.9 \\
Edit people/faces/character & 8.931 & 74.0 \\
Multi-step (compound) & 8.769 & 62.6 \\
Modify existing object & 8.663 & 62.0 \\
Alter background/texture/style & 8.579 & 56.9 \\
Cosplay special edit & 8.541 & 58.3 \\
Edit text or document content & 7.867 & 46.1 \\
Change camera view/perspective & 7.549 & 47.5 \\
\bottomrule
\end{tabular}
\end{table}

\subsubsection{Quantitative Results}
\label{subsubsection:quantitative}

\paragraph{ImgEdit-Bench results.}
\Cref{tab:imgedit-bench} reports the per-category and overall scores of BluePixel on ImgEdit-Bench, compared against a broad set of open-source and proprietary editing methods~\citep{brooks2023instructpix2pix, zhang2023magicbrush, zhao2024ultraedit, sheynin2024emu, cui2025emu35, flux2klein, flux2dev, batifol2025fluxkontext, qwenimage, seedream4, seedream5, nanobanana, gptimage}, ranging from early instruction-editing approaches to recent production-level models. BluePixel reaches an Overall score of 4.32, competitive with recent production-level models: it ranks first on Hybrid (4.59) and Action (4.93), tied for first on Add (4.93), and second on Adjust (4.76); the remaining gap to the strongest methods concentrates on Extract and Style.

\begin{table}[!t]
\centering
\caption{\bpxmodel ranks first on Hybrid and Action, ties for first on Add, and remains competitive overall. Main results on ImgEdit-Bench: each cell is the average score (1--5); Overall is the average across the nine task categories. Superscripts distinguish the reporting source for models with multiple reported scores: $^{a}$FireRed-Image-Edit, $^{b}$LongCat-Image, $^{c}$HiDream-O1-Image, $^{d}$Boogu-Image-0.1.}
\resizebox{\textwidth}{!}{%
\footnotesize\renewcommand{\arraystretch}{1.15}\begin{tabular}{lcccccccccc}
\toprule
Method & Add & Adjust & Extract & Replace & Remove & Background & Style & Hybrid (Compose) & Action & Overall \\
\midrule
GPT Image 2 & \underline{4.85} & \textbf{4.82} & \underline{4.07} & \textbf{4.92} & \textbf{4.75} & \textbf{4.88} & \underline{4.96} & \underline{4.47} & 4.85 & \textbf{4.73} \\
Qwen-Image-Edit-2511 & 4.54 & 4.57 & \textbf{4.13} & 4.70 & 4.46 & 4.36 & 4.89 & 4.16 & 4.81 & \underline{4.51} \\
Nano Banana 2.0 & 4.56 & 4.41 & 4.04 & 4.62 & \underline{4.74} & 4.54 & 4.87 & 4.01 & 4.81 & \underline{4.51} \\
Seedream 5.0 Lite & \textbf{4.93} & 4.69 & 3.01 & 4.41 & 4.45 & \underline{4.65} & 4.93 & 3.91 & 4.82 & 4.42 \\
Emu3.5 & 4.61 & 4.32 & 3.96 & \underline{4.84} & 4.58 & 4.35 & 4.79 & 3.69 & 4.57 & 4.41 \\
Nano Banana Pro & 4.44 & 4.62 & 3.42 & 4.60 & 4.63 & 4.32 & \textbf{4.97} & 3.64 & 4.69 & 4.37 \\
FLUX.2-Dev & 4.50 & 4.18 & 3.83 & 4.65 & 4.65 & 4.31 & 4.88 & 3.46 & 4.70 & 4.35 \\
Qwen-Image-Edit-2509$^{b}$ & 4.32 & 4.36 & 4.04 & 4.64 & 4.52 & 4.37 & 4.84 & 3.39 & 4.71 & 4.35 \\
Seedream 4.5 & 4.57 & 4.65 & 2.97 & 4.66 & 4.46 & 4.37 & 4.92 & 3.71 & 4.56 & 4.32 \\
Seedream 4.0$^{a,d}$ & 4.33 & 4.38 & 3.89 & 4.65 & 4.57 & 4.35 & 4.22 & 3.71 & 4.61 & 4.30 \\
Nano Banana$^{a,d}$ & 4.62 & 4.41 & 3.68 & 4.34 & 4.39 & 4.40 & 4.18 & 3.72 & 4.83 & 4.29 \\
Qwen-Image-Edit & 4.38 & 4.16 & 3.43 & 4.66 & 4.14 & 4.38 & 4.81 & 3.82 & 4.69 & 4.27 \\
GPT Image 1 [High] & 4.61 & 4.33 & 2.90 & 4.35 & 3.66 & 4.57 & 4.93 & 3.96 & \underline{4.89} & 4.20 \\
FLUX.1 Kontext [Pro]$^{b,c}$ & 4.25 & 4.15 & 2.35 & 4.56 & 3.57 & 4.26 & 4.57 & 3.68 & 4.63 & 4.00 \\
FLUX.1 Kontext [Dev]$^{a}$ & 3.99 & 3.88 & 2.19 & 4.27 & 3.13 & 3.98 & 4.51 & 3.23 & 4.18 & 3.71 \\
UltraEdit & 3.44 & 2.81 & 2.13 & 2.96 & 1.45 & 2.83 & 3.76 & 1.91 & 2.98 & 2.70 \\
MagicBrush & 2.84 & 1.58 & 1.51 & 1.97 & 1.58 & 1.75 & 2.38 & 1.62 & 1.22 & 1.90 \\
Instruct-Pix2Pix & 2.45 & 1.83 & 1.44 & 2.01 & 1.50 & 1.44 & 3.55 & 1.20 & 1.46 & 1.88 \\
\midrule
\rowcolor{bpxApp}\textbf{BluePixel} & \textbf{4.93} & \underline{4.76} & 3.45 & 4.53 & 4.21 & 4.43 & 3.85 & \textbf{4.59} & \textbf{4.93} & 4.32 \\
\bottomrule
\end{tabular}}
\label{tab:imgedit-bench}
\end{table}

\paragraph{VINS-4KEval results.}
We further evaluate the final model on the original VINS-4KEval benchmark at 4K resolution under the benchmark's default evaluation protocol (\Cref{tab:vin120k-x2edit}). This comparison includes FLUX.1 Kontext [Dev] + VINS, the Kontext-dev editor~\citep{batifol2025fluxkontext} post-adapted on VINS-4KEval~\citep{guan2026vins} and released alongside the benchmark, together with the open-source editors FLUX.1 Kontext [Dev], OmniGen2~\citep{wu2025omnigen2}, Step1X-Edit~\citep{liu2025step1xedit}, BAGEL~\citep{deng2025bagel}, ICEdit~\citep{zhang2025icedit}, and AnyEdit~\citep{yu2025anyedit}; our model matches the Kontext+VINS naturalness (4.70) and exceeds it on instruction adherence and detail.

\begin{table}[!t]
\centering
\caption{Quantitative comparison on the VINS-4KEval benchmark~\citep{guan2026vins} (509 tasks, 4K). BluePixel matches the naturalness of FLUX.1 Kontext [Dev] + VINS while exceeding it on instruction adherence and detail. ImgEditJudge scores each row on instruction following (Instr.\ Adh.) and naturalness (Nat.); Detail is their average. BluePixel is the final model after the full training recipe. Per-column best in bold, second best underlined.}
\label{tab:vin120k-x2edit}
\small
\setlength{\tabcolsep}{5pt}
\begin{tabular}{lccc}
\toprule
Method & Instr.\ Adh.$\uparrow$ & Nat.$\uparrow$ & Detail$\uparrow$ \\
\midrule
Nano Banana 2.0 & \textbf{4.68} & \textbf{4.79} & \textbf{4.74} \\
Seedream 4.0 & \underline{4.60} & \textbf{4.79} & \underline{4.70} \\
Seedream 5.0 Lite & 4.57 & 4.65 & 4.61 \\
\midrule
FLUX.1 Kontext [Dev] + VINS~\citep{guan2026vins} & 4.23 & \underline{4.70} & 4.47 \\
FLUX.1 Kontext [Dev] & 4.22 & 4.60 & 4.41 \\
OmniGen2 & 4.14 & 4.54 & 4.34 \\
Step1X-Edit & 4.06 & 4.50 & 4.28 \\
BAGEL & 4.15 & 4.39 & 4.27 \\
ICEdit & 3.76 & 4.42 & 4.09 \\
AnyEdit & 3.24 & 3.89 & 3.57 \\
\midrule
\textbf{BluePixel} & 4.52 & \underline{4.70} & 4.61 \\
\bottomrule
\end{tabular}
\end{table}

\paragraph{VINS-4KEval++ results.}
BluePixel achieves the second-highest Overall score (8.62) and success rate (63.2\%) among the seven models on VINS-4KEval++ (\Cref{tab:vins}). All models are evaluated on the 2{,}513 shared cases under the BluePixReward protocol. Per-metric, BluePixel ranks second in non-edited-region consistency (NC, 9.06), behind only GPT Image 2 (9.34), while instruction-following (IF, 8.74) and visual quality (VQ, 8.90) fall below four external models. Two caveats: (i)~the 0.02-point Overall margin over Nano Banana Pro is within noise given the absence of uncertainty estimates; (ii)~BluePixel applies prompt rewriting, whereas each external model produces a single output per case.

\begin{table}[!t]
\centering
\caption{BluePixel ranks second on VINS-4KEval++ behind GPT Image 2, with the second-best non-edited-region consistency: BluePixReward scores on the 2{,}513 cases shared by all models. BluePixel reports the mean of an eight-seed run with prompt rewriting (per-seed detail in \Cref{tab:seed}); each external model has one output per case. Scores are on a 0--10 scale, with higher values indicating better performance. Success is the percentage of outputs judged successful by the same evaluator.}
\label{tab:vins}
\small
\setlength{\tabcolsep}{4pt}
\begin{tabular}{lrrrrr}
\toprule
Method & Overall$\uparrow$ & IF$\uparrow$ & NC$\uparrow$ & VQ$\uparrow$ & Success (\%)$\uparrow$ \\
\midrule
GPT Image 2~\citep{gptimage} & 9.15 & 9.22 & 9.34 & 9.42 & 74.8 \\
Nano Banana Pro~\citep{nanobanana} & 8.60 & 8.80 & 8.72 & 9.12 & 59.4 \\
Nano Banana 2.0~\citep{nanobanana} & 8.56 & 8.81 & 8.63 & 9.08 & 57.9 \\
Wan 2.7 Pro~\citep{wan27pro} & 8.56 & 8.72 & 8.77 & 9.04 & 59.5 \\
Seedream 5.0 Lite~\citep{seedream5} & 8.44 & 8.95 & 8.53 & 8.81 & 51.8 \\
Kling~\citep{klingai} & 7.90 & 8.01 & 8.37 & 8.55 & 45.4 \\
\midrule
\textbf{BluePixel} & 8.62 & 8.74 & 9.06 & 8.90 & 65.1 \\
\bottomrule
\end{tabular}
\end{table}

%% ================================================================
%%  CONCLUSION
%% ================================================================

\section{Conclusion}
\label{section:conclusion}

We have presented \bpxmodel, a native 4K image editing model with an agent framework, and a training recipe that makes native-4K editing practical on limited resources: progressive VAE adaptation through $f{:}\,8 \to 32 \to 64$ yields an efficient 4K initialization, group-wise ACE-LoRA continual learning with MOPD consolidation absorbs a long tail of vertical editing skills with limited forgetting, and reinforcement learning with the purpose-trained BluePixReward refines the aesthetic--fidelity trade-off. Distillation via \emph{source-anchored target projection} (SATP) then compresses the 50-step teacher to 4 steps, completing a 4K edit in 8.7\,s on a single Kunlunxin P800 XPU while preserving task success rate and background consistency. BluePixel Agent wraps the model in a closed loop that decomposes requests across specialist and general editing tools, grounds underspecified instructions through retrieval and prompt rewriting, and verifies every candidate with bounded recovery. A unified data pipeline (a VLM-based quality gate and a four-dimensional reward model) provides one assessment basis from training-data filtering to final evaluation. On VINS-4KEval++, BluePixel ranks second only to GPT Image 2 in overall score among seven models, and second in non-edited-region consistency. Together, these results suggest a path beyond the single-model paradigm that currently dominates image editing research.

\clearpage
\setlength{\bibsep}{3pt}
\bibliographystyle{ref/assets/plainnat}
\bibliography{ref/paper}

\clearpage
\beginappendix

\section*{The BluePixel Team}

BluePixel was developed through the collective efforts of the following team members and collaborators.

\paragraph{Research \& Algorithm.}
Shanyan Guan led the BluePixel project and designed the overall framework, core algorithms, data construction, and evaluation pipeline. Shiyu Li developed the BluePixel Agent harness. Xin Li contributed to data construction, training instruction refinement, and instruction recommendation, and developed the reward model for editing evaluation. Zhizhou Chen validated the initial training algorithm. Yuehao Liu developed the continual learning algorithm for acquiring multiple editing capabilities. Yunzhe Xu developed the distillation pipeline. Ruixue Shen developed the data construction pipeline and explored DPO-based post-training.

Chao Ma (Shanghai Jiao Tong University) and Ying Tai (Nanjing University) contributed as long-term research collaborators. Chao Ma, Ying Tai, Zhongang Qi, Wei Li, and Yanhao Ge provided valuable guidance and support throughout the project.

\paragraph{Engineering.}
Min Jin, Junbo Liu, Qiang Liu, Penggang Shi, Sheng Sun, Xiao Wang, and Yong Zhang developed and maintained the training and deployment infrastructure. Shiju Zhang contributed to training data construction. Names are listed alphabetically by surname.

\section{Hairstyle Sub-templates}
\label{app:hairstyle_templates}

\Cref{tab:hairstyle_templates} lists the 26 hairstyle sub-templates used in the continual learning experiments.

\begin{table}[h]
    \centering
    \small
    \caption{Full list of hairstyle sub-templates for the Change Hairstyle task.}
    \label{tab:hairstyle_templates}
    \begin{tabular}{rl|rl|rl}
    \toprule
    \# & Template & \# & Template & \# & Template \\
    \midrule
    1 & side-part & 2 & braids & 3 & bangs \\
    4 & half-up princess & 5 & half-up bun & 6 & curly \\
    7 & glowing strands & 8 & flying hair & 9 & short wavy \\
    10 & layered perm & 11 & long wavy & 12 & latin curl \\
    13 & side bangs & 14 & japanese short curly & 15 & japanese bangs \\
    16 & straight & 17 & sakura & 18 & french bangs \\
    19 & above-brown bangs & 20 & air bangs & 21 & elf short \\
    22 & textured middle & 23 & wool roll & 24 & collar-length \\
    25 & flowing ponytail & 26 & fishtail perm & & \\
    \bottomrule
    \end{tabular}
\end{table}

\end{document}